\documentclass[letterpaper, 10 pt, conference]{ieeeconf}  

\IEEEoverridecommandlockouts                              

\usepackage{kotex} 
\usepackage{graphicx}
\usepackage{booktabs}
\usepackage{xcolor}
\usepackage{subcaption}
\usepackage[hidelinks,
  pdfauthor={YoungJae Cheong and Jhonghyun An},
  pdftitle={Source-Only Cross-Weather LiDAR via Geometry-Aware Point Drop}]{hyperref}
\usepackage{amssymb}
\usepackage{multirow}
\usepackage{makecell}
\usepackage{amsmath,amssymb}
\usepackage{algorithm}          
\usepackage{algpseudocode}      


\title{\LARGE \bf
Source-Only Cross-Weather LiDAR via Geometry-Aware Point Drop}

\author{YoungJae Cheong, Jhonghyun An\textsuperscript{*}%
\thanks{This work was supported by the Project for Collaboration R\&D between Industry, University, and Research Institute, funded by the Ministry of SMEs and Startups of Korea in 2025 (RS-2025-02220569).}%
\thanks{YoungJae Cheong and Jhonghyun An are with the School of Computing, Gachon University, Seongnam, Republic of Korea (e-mail: bluebull777@gachon.ac.kr; jhonghyun@gachon.ac.kr). Corresponding author: Jhonghyun An.}%
}

\begin{document}

\maketitle
\thispagestyle{empty}
\pagestyle{empty}

\begin{abstract}
Adverse weather conditions, such as rain, snow, and fog, severely degrade LiDAR semantic segmentation by introducing refraction, scattering, and point dropouts that compromise geometric integrity. While prior approaches ranging from weather simulation and mixing-based augmentation to domain randomization and regularization enhance robustness, they frequently overlook structural vulnerabilities inherent to object boundaries, corners, and highly sparse regions.

To address this limitation, we propose a Light Geometry-Aware Adapter. This module aligns azimuths and applies horizontal circular padding to preserve neighbor continuity across the $0^\circ$--$360^\circ$ wrap-around boundary. Using a local-window K-Nearest Neighbors (KNN) search, it aggregates nearby points and computes lightweight local statistics, compressing them into compact geometry-aware cues. During training, these cues facilitate region-aware regularization, which effectively stabilizes predictions in structurally fragile areas. The proposed adapter is designed to be plug-and-play, complements existing augmentation techniques, and operates exclusively during training, incurring negligible inference overhead.

Operating under a rigorous \emph{source-only} cross-weather paradigm wherein models are trained on SemanticKITTI and evaluated on SemanticSTF without target-domain labels or fine-tuning, our adapter achieves a +3.4 mIoU improvement over strong data-centric augmentation baselines. Furthermore, it demonstrates performance comparable to advanced class-centric regularization methods. These findings highlight that geometry-driven regularization constitutes a critical pathway toward achieving highly robust, all-weather LiDAR segmentation.
\end{abstract}

\section{INTRODUCTION}
LiDAR provides accurate ranging for autonomous driving, but adverse weather such as rain, snow, and fog induces refraction, scattering, and point dropouts that distort point clouds and degrade segmentation~\cite{park2025ntn,park2024lidarweather}.
\emph{Foreground} classes (vehicles, pedestrians, riders) are especially vulnerable due to low point density, making errors \textbf{safety-critical}.
Generic regularizers such as Dropout~\cite{srivastava2014dropout} do not directly address \emph{structural} distortions, including boundary breaks, corner loss, and noisy sparse regions.
We study \emph{source-only} cross-weather transfer, training on SemanticKITTI~\cite{behley2019semantickitti} and evaluating on SemanticSTF~\cite{xiao2023pointdr} without target labels or fine-tuning.

Prior work largely follows two strategies.
The first uses data augmentation and \emph{domain randomization} to mimic weather via dropping, jittering, or simulation~\cite{xiao2023pointdr,xiao2022polarmix,zhao2024unimix}.
The second encourages \emph{invariance and consistency} under distribution shifts through mixing-based generalization, uncertainty mitigation, or regularization.
However, augmentation alone often fails to capture \textbf{structural vulnerabilities} at boundaries and in sparse regions, and consistency-based methods frequently omit explicit local geometry, which can increase confusion between boundary-adjacent classes.
We adopt the source-only setting throughout.

\begin{figure}[t]
  \centering
  \includegraphics[width=\columnwidth]{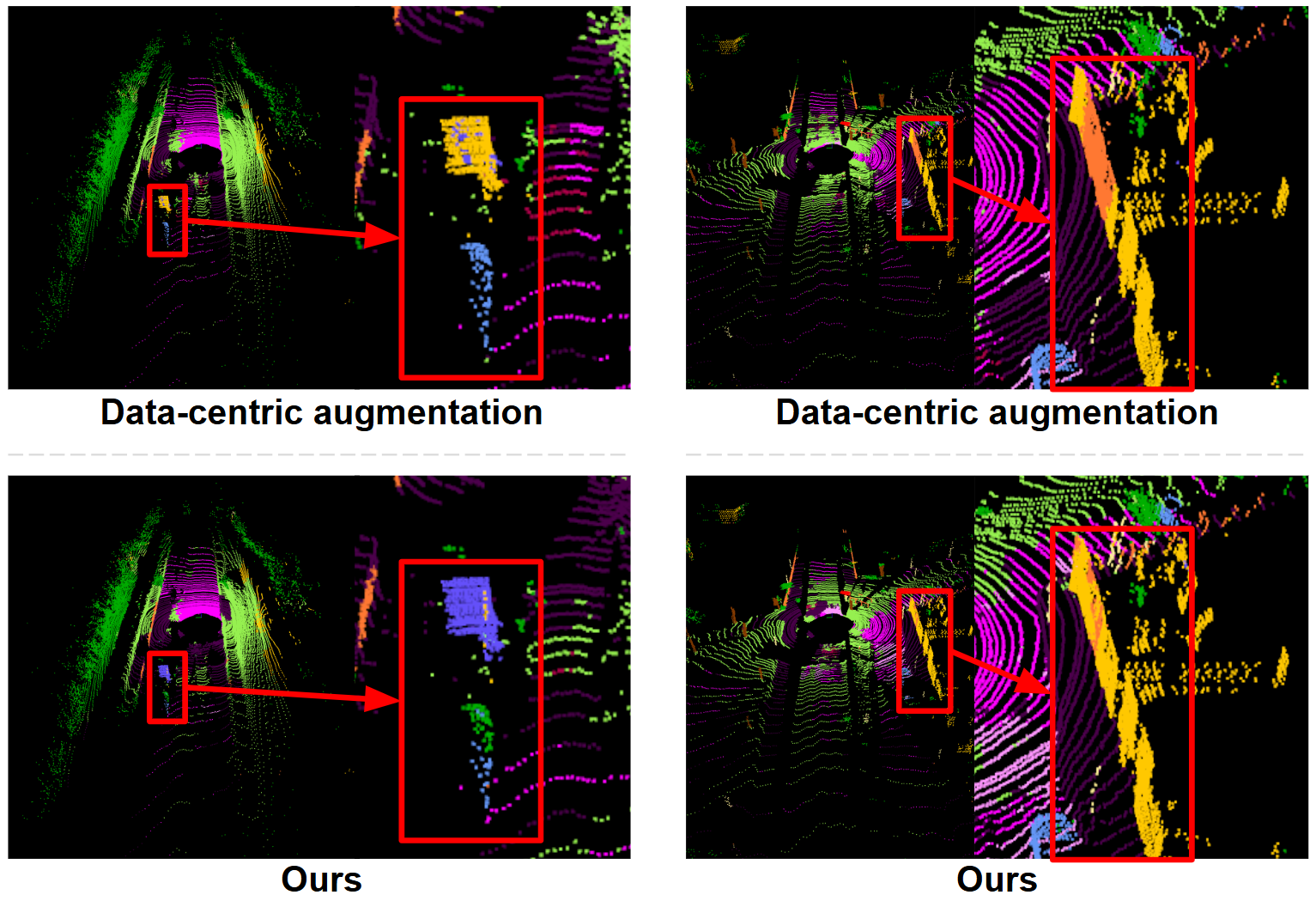}
  \caption{\textbf{Failure Cases in Adverse Weather}~\cite{park2024lidarweather}. (a) The baseline misclassifies a bus, merging it into a nearby building. Conversely, our \emph{Light Geometry-Aware Adapter} preserves local neighbor continuity, accurately segmenting the bus (blue). (b) The baseline blurs the semantic boundary between a building and a fence, whereas our method distinctly preserves this structural boundary.}
  \label{fig:limitations_Data-centric augmentation}
\end{figure}

Fig.~\ref{fig:limitations_Data-centric augmentation} elucidates the core motivation of our approach. A standard data-centric augmentation baseline~\cite{park2024lidarweather} is prone to merging a \textit{bus} into an adjacent \textit{building} or blurring the precise boundary between a \textit{building} and a \textit{fence} when neighbor relations are corrupted by weather noise. In contrast, our proposed methodology actively preserves neighbor continuity and curtails boundary leakage by explicitly modeling local point-to-point structural relationships.

\begin{figure}[t]
  \centering
  \includegraphics[width=\columnwidth]{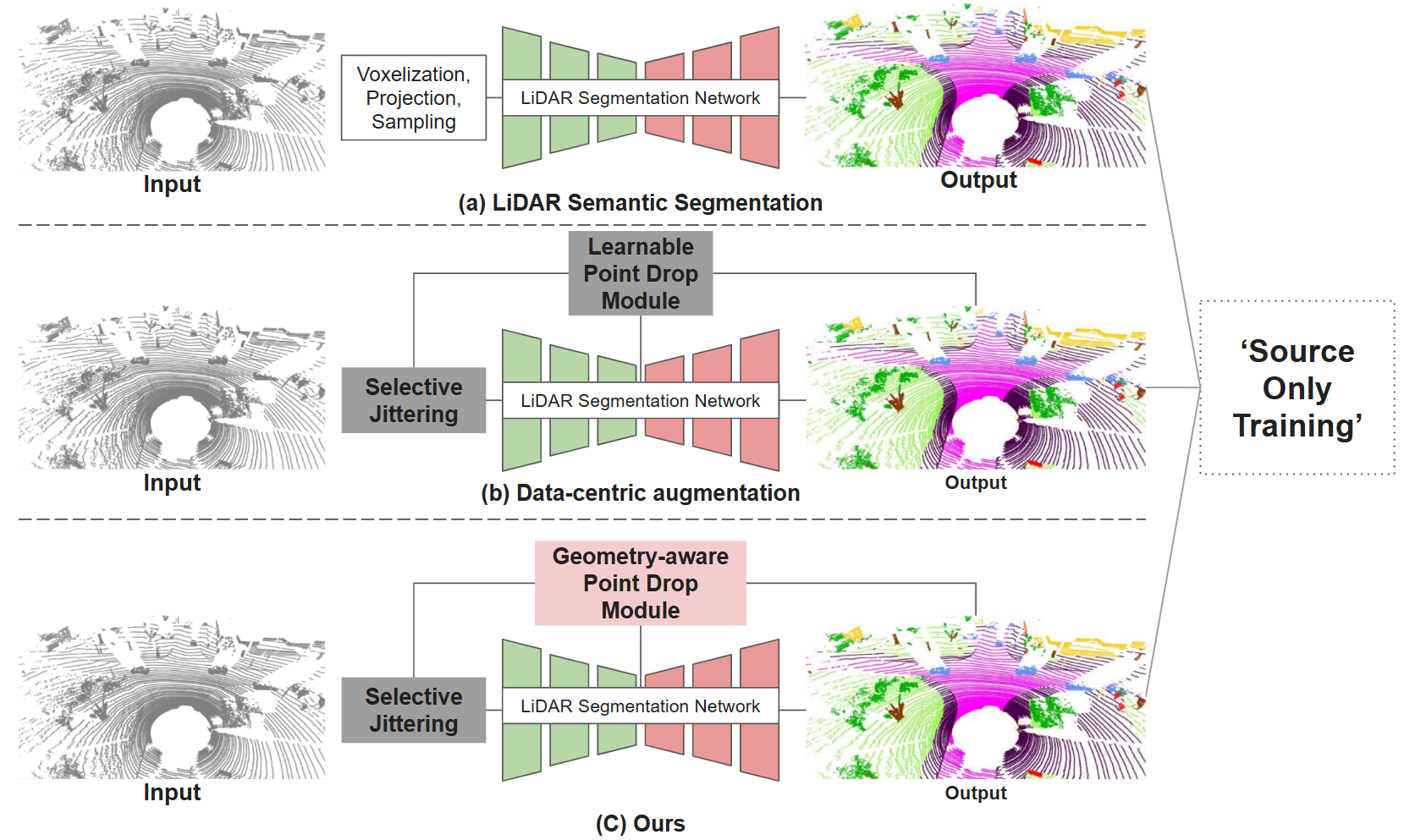}
  \caption{Comparison of LiDAR segmentation pipelines under the \emph{source-only} setting. (a) Conventional networks lacking structural noise modeling. (b) Data-centric augmentation employing selective jittering and learnable point drop~\cite{park2024lidarweather}. (c) Our pipeline, featuring the \emph{Light Geometry-Aware Adapter}, which systematically enhances region-level dropping by anchoring it to local geometric continuity.}
  \label{fig:architecture}
\end{figure}

\begin{figure*}[t]
  \centering
  \includegraphics[width=\textwidth]{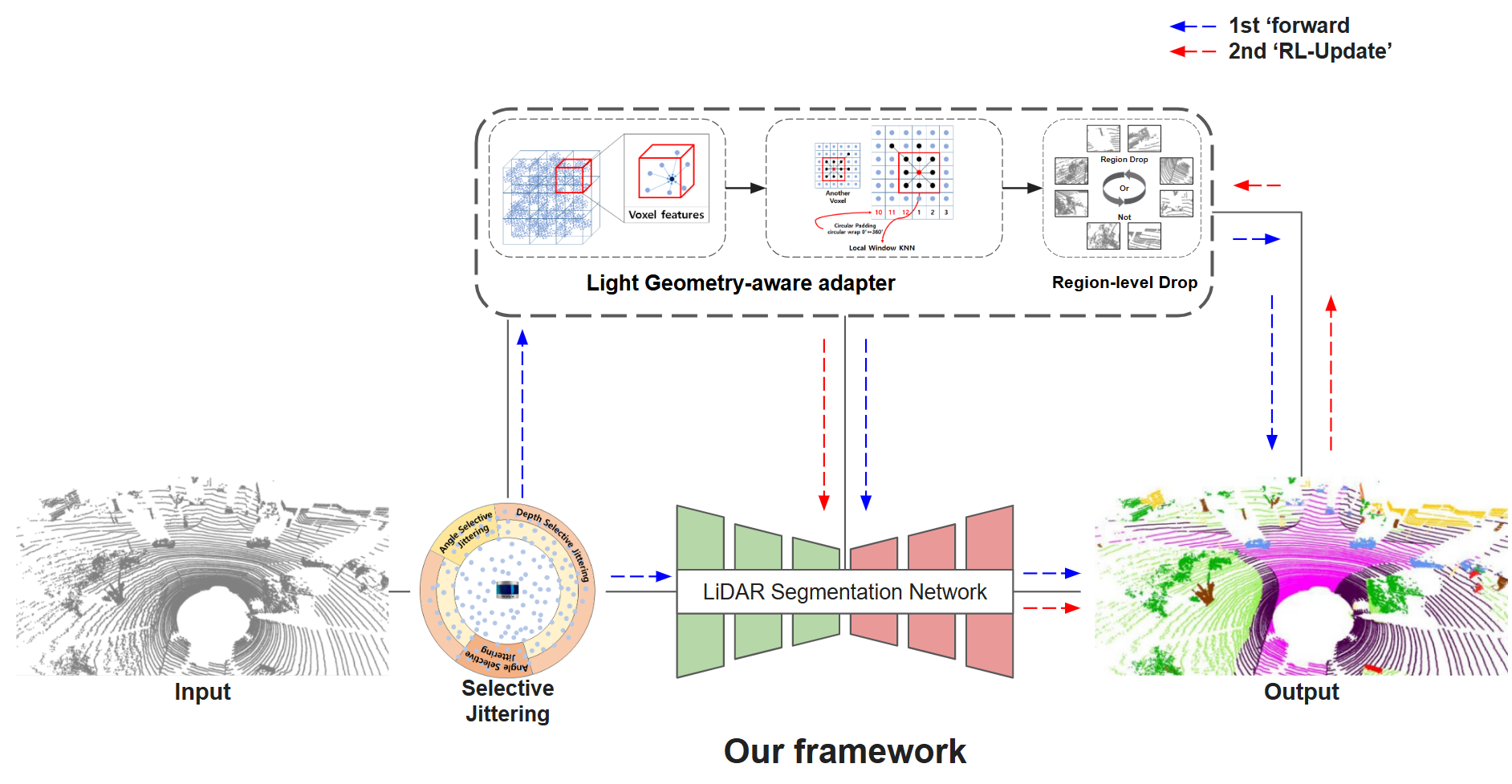}
    \caption{The Overall Pipeline. Selective jittering emulates weather-induced artifacts~\cite{park2024lidarweather}. Before the learnable point drop module, the \emph{Light Geometry-Aware Adapter} extracts compact geometry-aware cues via local-window KNN with circular wrapping to preserve $0^\circ$--$360^\circ$ continuity. These cues guide \textbf{region-level regularization} at boundaries, corners, and sparse structures, improving robustness under adverse weather.}
  \label{fig:overall}
\end{figure*}

\begin{figure}[t]
  \centering
  \includegraphics[width=\columnwidth]{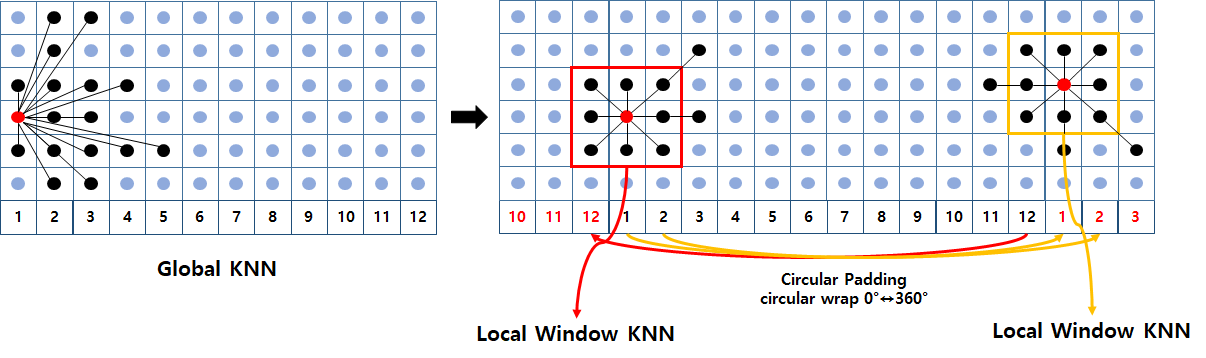}
    \caption{Global KNN constructs neighborhoods from the entire scan, often matching distant points across the azimuth seam, which increases overhead and weakens locality~\cite{zhao2024triplemixer}. In contrast, our \emph{Light Geometry-Aware Adapter} uses horizontal local-window KNN with circular padding to wrap the $0^\circ$--$360^\circ$ seam, preserving edge and corner cues with lightweight computation.}
  \label{fig:Local-window KNN}
\end{figure}

We therefore introduce a \textbf{Light Geometry-aware adapter} that injects two geometry signals into training.
First, it preserves \emph{neighbor continuity} from the LiDAR scanning pattern by aligning azimuth and applying horizontal circular padding across the $0^\circ$--$360^\circ$ seam.
Second, it aggregates a \emph{local-window KNN} neighborhood to compute simple local statistics (offsets and dispersion) that summarize \emph{structural risk} at boundaries, corners, and sparse zones.
These geometry-aware cues guide \emph{region-aware regularization} and stabilize learning under weather-induced sparsity.
Unlike PointDR~\cite{xiao2023pointdr}, which broadens the training distribution via large-scale domain randomization, our adapter explicitly links local geometry to the learning signal.
It is lightweight, plug-and-play, and complementary to augmentation.
On SemanticKITTI$\rightarrow$SemanticSTF, we obtain consistent gains over strong data-centric baselines and also improve over a class-centric regularization baseline~\cite{park2025ntn}.
Fig.~\ref{fig:architecture} illustrates how the adapter enhances region-level dropping for robust segmentation in adverse weather.

\noindent\textbf{Our main contributions are:}
\begin{enumerate}
\item \textbf{Light Geometry-aware adapter.} We distill a lightweight geometry module that preserves neighbor continuity via circular padding and local-window KNN, providing geometry-aware cues before dropping with low computation.
\item \textbf{Geometry-conditioned robustness.} We introduce region-aware regularization driven by these cues to reduce confusion at boundaries, corners, and sparse regions under adverse weather.
\end{enumerate}

\section{Related Work}

\subsection{Semantic Segmentation of LiDAR Point Clouds}
LiDAR semantic segmentation methods are typically categorized into \emph{voxel-based}, \emph{projection-based (range-view)}, and \emph{point-based} approaches.
Voxel-based models exploit sparse 3D convolutions for large-scale outdoor scenes, e.g., SPVCNN~\cite{tang2020spvcnn} and Cylinder3D~\cite{zhou2020cylinder3d}, and transformer variants further extend this line, e.g., SphereFormer~\cite{lai2023sphereformer}.
Projection-based methods render scans to range images and apply 2D CNN backbones, such as SqueezeSeg~\cite{wu2018squeezeseg} and RangeNet++~\cite{milioto2019rangenetpp}.
Point-based models operate directly on points, from PointNet/PointNet++~\cite{qi2017pointnet,qi2017pointnetpp} to kernel-based and lightweight designs, e.g., KPConv~\cite{thomas2019kpconv} and RandLA-Net~\cite{hu2020randlanet}.
While accuracy has improved steadily, adverse-weather robustness remains challenging, motivating geometry-aware components that stabilize local structure.

\subsection{Augmentation and Domain Generalization for LiDAR}
Data augmentation and synthesis broaden training distributions through geometric and intensity perturbations, as well as mixing-based strategies.
PolarMix~\cite{xiao2022polarmix} and UniMix~\cite{zhao2024unimix} recombine scans to increase diversity and reduce domain gaps.
Domain generalization and adaptation further align representations with distribution-matching objectives, e.g., DGAF-L~\cite{li2018dgafl} and PCL~\cite{yao2022pcl}.
These methods improve coverage or alignment but do not explicitly preserve neighborhood continuity under weather-induced sparsity.
Our adapter is complementary, providing local geometry cues that can be used alongside existing augmentations.

\subsection{LiDAR Perception in Adverse Weather}
Environmental phenomena such as rain, snow, and fog introduce severe refraction, scattering, and point dropouts, precipitously degrading perception accuracy~\cite{park2025ntn,park2024lidarweather}. The SemanticSTF dataset~\cite{xiao2023pointdr} provides a critical real-world benchmark for evaluating these specific failure modes. Recent data-centric pipelines have integrated selective jittering with Reinforcement Learning (RL)-guided point dropping to systematically target noise-driven vulnerabilities~\cite{park2024lidarweather}. In a related vein, TripleMixer~\cite{zhao2024triplemixer} employs a Geometry Mixer to model local relations. Distinct from these approaches, our \emph{Light Geometry-Aware Adapter} extracts non-parametric geometric cues from local-window neighborhoods utilizing circular wrapping. These specific cues directly inform and support region-aware training objectives, dynamically adapting to structural fragility in adverse weather.

\section{Method}

\subsection{Overview}
Our objective is to strengthen a \emph{data-centric augmentation} pipeline based on Selective Jittering (SJ) and an RL-based Learnable Point Drop (LPD) module~\cite{park2024lidarweather} by integrating a \textbf{Light Geometry-Aware Adapter}, so that \emph{structural vulnerabilities} under adverse weather are explicitly reflected in the training signal (Fig.~\ref{fig:overall}). SJ emulates weather-induced corruptions, while LPD selects dropping regions and ratios from region-level statistics. The adapter is inserted \textbf{only before} LPD. It preserves neighborhood continuity across the $0^\circ$--$360^\circ$ seam via horizontal \emph{circular wrapping/padding} and extracts compact \emph{geometry-aware cues} using a \emph{local-window KNN}. These cues are injected into the RL agent state to steer region-level dropping toward boundaries, corners, and sparse regions, reducing boundary-adjacent semantic confusion (e.g., \textit{bus}$\leftrightarrow$\textit{building}; Fig.~\ref{fig:limitations_Data-centric augmentation}). Unlike PointDR~\cite{xiao2023pointdr}, which relies on domain randomization, our adapter deterministically couples local geometry with the dropping policy. We evaluate this architecture under the SemanticKITTI$\rightarrow$SemanticSTF \emph{source-only} transfer setting.

\subsection{In Detail}

\textbf{Selective Jittering (SJ) :}
We adopt SJ as introduced in prior \emph{data-centric augmentation} work~\cite{park2024lidarweather}.
SJ injects \emph{non-uniform}, frame-wise perturbations that mimic adverse weather by selectively applying small offsets or noise to a subset of points, conditioned on simple factors such as range, bearing, and intensity.
Unperturbed points remain unchanged.
The perturbed input is fed to both the reference and drop branches.
This produces supervision that captures \emph{noise-driven} vulnerabilities alongside \emph{structure-driven} ones.

\textbf{Light Geometry-aware adapter.}
For each query point $p_i$, we retrieve $K$ neighbors $\mathcal{N}_K(i)=\{p_i^1,\ldots,p_i^K\}$ using a horizontal \emph{local-window KNN} and apply \emph{circular padding} to preserve neighbor continuity across the $0^\circ$--$360^\circ$ azimuth seam.
Compared to conventional mixers that use a \emph{global KNN} search over the entire scan~\cite{zhao2024triplemixer}, our restricted candidate window reduces spurious cross-boundary matches and lowers computation (Fig.~\ref{fig:Local-window KNN}).
We first stabilize local features by voxel-wise mean pooling, yielding a base feature $v_i$ for $p_i$ within the window, and compute a windowed mean coordinate $\mu_i$.
We then summarize local geometry with two lightweight dispersion indicators:
$d_i^{(1)}=\lVert p_i-\mu_i\rVert$ (off-center geometry near boundaries/corners) and
$d_i^{(2)}=\frac{1}{K}\sum_{k=1}^{K}\lVert p_i^k-\mu_i\rVert$ (spread/compactness of local support).
Using these quantities, we mix $p_i$ with each neighbor and pool them with attention:
\begin{equation}\label{eq:lga_lk}
\ell_i^{k}=\phi_{p}\!\big(p_i \oplus p_i^{k} \oplus (p_i-p_i^{k}) \oplus \mu_i \oplus [\,d^{(1)}_i,d^{(2)}_i\,]\big),
\end{equation}
\begin{gather}
\alpha_i^{k} = \operatorname{FC}(\ell_i^{k}),\qquad
c_i^{k} = \operatorname{Softmax}_{k}(\alpha_i^{k}) \notag\\
f_i^{\mathrm{pt}} = \sum_{k=1}^{K} c_i^{k}\,\ell_i^{k}. \label{eq:lga_att_pool}
\end{gather}
Finally, we form a geometry-aware cue $g_i$ by fusing the pooled feature $f_i^{\mathrm{pt}}$ with the base feature $v_i$:
\begin{equation}\label{eq:lga_out}
g_i=\operatorname{MLP}\!\big(v_i \oplus f_i^{\mathrm{pt}}\big)\;\oplus\; p_i ,
\end{equation}
where $g_i$ augments only the \emph{pre-LPD} agent state, keeping the inference path unchanged.
$g_i$ is the final geometry-aware cue.

\begin{table*}[t]
\centering
\scriptsize
\setlength{\tabcolsep}{2.5pt}
\caption{Quantitative evaluation on SemanticKITTI$\rightarrow$SemanticSTF. Our method improves per-class IoU and overall mIoU over strong baselines, showing that geometry-conditioned, region-aware point dropping enhances structural robustness for cross-weather adaptation.}
\label{tab:semantic_kitti_to_stf_final}
\begin{tabular}{l|ccccccccccccccccccc|c}
\toprule
\textbf{Method} & car & bi.cle & mt.cle & truck & bus & pers & bi.clst & mt.clst & road & parki. & sidew. & oth.g. & build. & fence & veget. & trunk & terra. & pole & traf. & \textbf{mIoU} \\
\midrule
Oracle                & 89.4 & 42.1 & 0.0  & 59.9 & 61.2 & 69.6 & 39.0 & 0.0  & 82.2 & 21.5 & 58.2 & 45.6 & 86.1 & 63.6 & 80.2 & 52.0 & 77.6 & 50.1 & 61.7 & 54.7 \\
Source-only           & 55.9 & 0.0  & 0.2  & 0.2  & 10.9 & 10.3 & 6.0  & 0.0  & 61.2 & 10.9 & 32.0 & 0.0  & 67.9 & 41.6 & 49.8 & 27.9 & 40.8 & 29.6 & 15.7 & 24.4 \\
Dropout~\cite{srivastava2014dropout} & 62.1 & 0.0  & 15.5 & 3.0  & 11.5 & 5.4  & 2.0  & 0.0  & 58.4 & 12.8 & 26.7 & 1.1  & 72.1 & 43.6 & 52.9 & 34.2 & 43.5 & 28.4 & 15.5 & 25.7 \\
Perturbation          & 74.4 & 0.0 & 0.0 & 23.3 & 0.6  & 19.7 & 20.0 & 0.0  & 0.0  & 59.3 & 10.7 & 32.0 & 7.2  & 70.2 & 45.2 & 57.1 & \underline{47.9} & 28.2 & 16.2 & 25.9 \\
PolarMix~\cite{xiao2022polarmix}     & 57.8 & 1.8  & 3.6  & 3.7  & \underline{26.5} & 3.7  & \underline{26.5} & 0.0  & 65.7 & 2.9  & 35.9 & 48.7 & 71.0 & \underline{58.7} & 53.8 & 20.5 & 45.4 & 29.3 & 15.8 & 26.6 \\
MMD~\cite{li2018dgafl}          & 63.6 & 0.0  & 2.6  & 17.4 & 11.4 & 28.1 & 0.0  & 0.0  & 67.0 & 14.1 & 37.6 & 41.2 & 67.1 & 41.2 & 57.1 & 22.4 & 47.9 & 28.2 & 16.2 & 26.9 \\
PCL~\cite{yao2022pcl}               & 65.9 & 0.0  & 0.4  & 17.3 & 8.4  & 8.4  & 8.4  & 0.0  & 59.6 & 12.0 & 35.3 & \underline{63.1} & \underline{74.0} & 47.5 & \underline{60.7} & 15.8 & 48.9 & 26.1 & 27.5 & 26.4 \\
PointDR~\cite{xiao2023pointdr}            & 67.3 & 0.0  & 4.5  & 19.9 & 18.8 & 2.7  & 20.0 & 0.0  & 62.6 & 12.9 & 36.8 & 43.8 & 73.3 & 43.8 & 56.4 & 32.2 & 45.7 & 28.7 & 27.4 & 26.4 \\
DGLSS~\cite{kim2023dglss}            & 72.6 & 0.1  & 11.7 & 29.4 & 13.7 & 48.3 & 0.5  & 21.2 & 65.0 & \underline{20.2} & 36.5 & 3.8  & \underline{78.9} & 51.8 & 57.0 & 36.4 & 42.7 & 26.9 & 34.9 & 34.6 \\
UniMix~\cite{zhao2024unimix}         & 82.7 & \underline{6.6}  & 8.6  & 4.5  & 19.9 & 35.5 & 15.1 & 15.5 & 55.8 & 10.2 & 36.5 & 40.1 & 72.8 & 40.1 & 49.1 & 23.5 & 39.4 & 23.5 & 31.5 & 31.5 \\
DGUIL~\cite{he2024dguil}             & 77.9 & 1.0  & 19.1 & 26.0 & 9.7  & 46.3 & 0.6  & 9.3  & \underline{69.1} & 9.8  & 38.6 & 9.4  & 73.3 & 51.2 & 59.0 & 31.8 & \underline{50.8} & 31.8 & 22.3 & 31.4 \\
\midrule
LiDARWeather~\cite{park2024lidarweather} & 83.1 & 1.2  & 17.2 & 30.5 & 18.4 & 47.5 & 1.07 & 18.8 & 64.0 & 15.9 & \underline{38.7} & 4.6  & 77.4 & 50.8 & 59.7 & 37.2 & 47.7 & 31.1 & 35.8 & 35.8 \\
No Thing, Nothing~\cite{park2025ntn} & \underline{83.3} & 3.7  & \underline{31.3} & \underline{36.2} & 18.2 & \underline{53.3} & 6.8  & \underline{45.9} & 67.2 & 18.1 & 37.2 & 5.4  & 72.1 & 41.8 & 58.0 & 36.0 & 46.0 & \underline{39.8} & \underline{38.9} & \underline{38.9} \\
Ours & \textbf{85.0} & \textbf{9.9} & 23.9 & \textbf{38.7} & 23.6 & 46.3 & 9.7 & 28.8 & 65.4 & 13.1 & 37.5 & 1.8 & \textbf{77.8} & 50.6 & \textbf{65.2} & \textbf{39.4} & \textbf{53.8} & 32.7 & \textbf{40.7} & \textbf{39.2} \\
\bottomrule
\end{tabular}
\end{table*}

\begin{algorithm}[t]
\caption{LPD with Light Geometry-aware Adapter}
\label{alg:lpd_gmx_Light}
\begin{algorithmic}[1]
\Require input cloud $P$, backbone $f$, decision module $Q$, drop ratios $B$, selective jittering $SJ$, Light Geometry-aware Adapter $adapter$
\Ensure perturbed cloud $P'$, geometry-aware cues $g$, region features $\{F_k\}$
\State $\tilde P \gets \mathrm{SJ}(P)$; \quad $u \gets \mathrm{Uncertainty}(f(\tilde P))$
\State $g \gets \mathrm{adapter}(\tilde P)$ \quad 
\State $\{(\mathcal{R}_k,F_k)\} \gets \mathrm{BuildRegions}(\tilde P, g)$ \quad 
\State $(k^\star,b^\star) \gets \varepsilon\text{-greedy } \arg\max_{k,b} Q\big([F_k,u],\, b\big)$ \quad 
\State $P' \gets \mathrm{RegionDrop}\!\left(\tilde P, \mathcal{R}_{k^\star}, B[b^\star]\right)$
\State $r \gets \mathrm{SegLoss}(f(P')) - \mathrm{SegLoss}(f(\tilde P)) - \lambda \cdot \mathrm{gt\_ratio}$
\State update $Q$ with $\big(s=[F_{k^\star},u],\, a=(k^\star,b^\star),\, r,\, s'\big)$; \quad update $f$
\end{algorithmic}
\end{algorithm}

\textbf{RL-Based Region-Level Point Drop:}
We retain the core LPD architecture established in prior work~\cite{park2024lidarweather}, purposefully \emph{augmenting} the RL agent's state representation with the newly derived geometry-aware cues. While the standard LPD primarily identifies \emph{noise-induced} vulnerabilities via rudimentary loss and uncertainty metrics, our enhanced state matrix provides an explicit, quantifiable signal for \emph{structural} vulnerabilities at geometric boundaries, corners, and sparse zones. This geometrically-grounded state guides the DQN policy~\cite{mnih2013dqn} to selectively target regions exhibiting high structural risk, effectively coupling reinforcement learning decisions with concrete geometric evidence.

\textbf{Overall Workflow:}
Algorithm~\ref{alg:lpd_gmx_Light} details the integrated procedure. Given an SJ-perturbed input $\tilde P$, the adapter extracts geometry-aware cues $\{g_i\}$ and constructs region descriptors $\{(\mathcal{R}_k,F_k)\}$ prior to the LPD phase. Specifically, \textsc{BuildRegions} groups points into candidate regions via a coarse spatial partitioning (e.g., voxelization with size $\Delta$), where each region $\mathcal{R}_k$ is the set of point indices falling into the same cell. For each region, we compute the region-level descriptor $F_k$ by applying a permutation-invariant aggregation (e.g., mean pooling) over point-level cues $\{g_i \mid i \in \mathcal{R}_k\}$, optionally concatenated with simple region statistics such as point count or density. These descriptors are concatenated with the segmentation backbone's initial predictions and calculated uncertainty to construct the Q-network state matrix. An $\varepsilon$-greedy policy subsequently dictates the target drop regions and ratios.

\subsection{Loss Formulation}
Following the established dual-branch paradigm~\cite{park2024lidarweather}, the reference branch operates solely with selective jittering (SJ), while the drop branch executes the LPD mechanism informed by our Light Geometry-Aware Adapter. Let $i\in\{1,\dots,N\}$ and $c\in\{1,\dots,C\}$ denote point and class indices, respectively. Ground-truth labels are defined as $Y=\{y_i^{(c)}\}$, with predicted probabilities $\hat Y=\{\hat y_i^{(c)}\}$ and corresponding class weights $w_c$. The adapter outputs a normalized vulnerability score $s_i\in[0,1]$ for each point $i$. We utilize $\kappa\ge0$ for geometry-aware loss reweighting, alongside $\alpha\ge1$ and $\eta\ge0$ to balance the post-drop and entropy regularization terms.

\begin{figure*}[t]
  \centering
  \includegraphics[width=\textwidth]{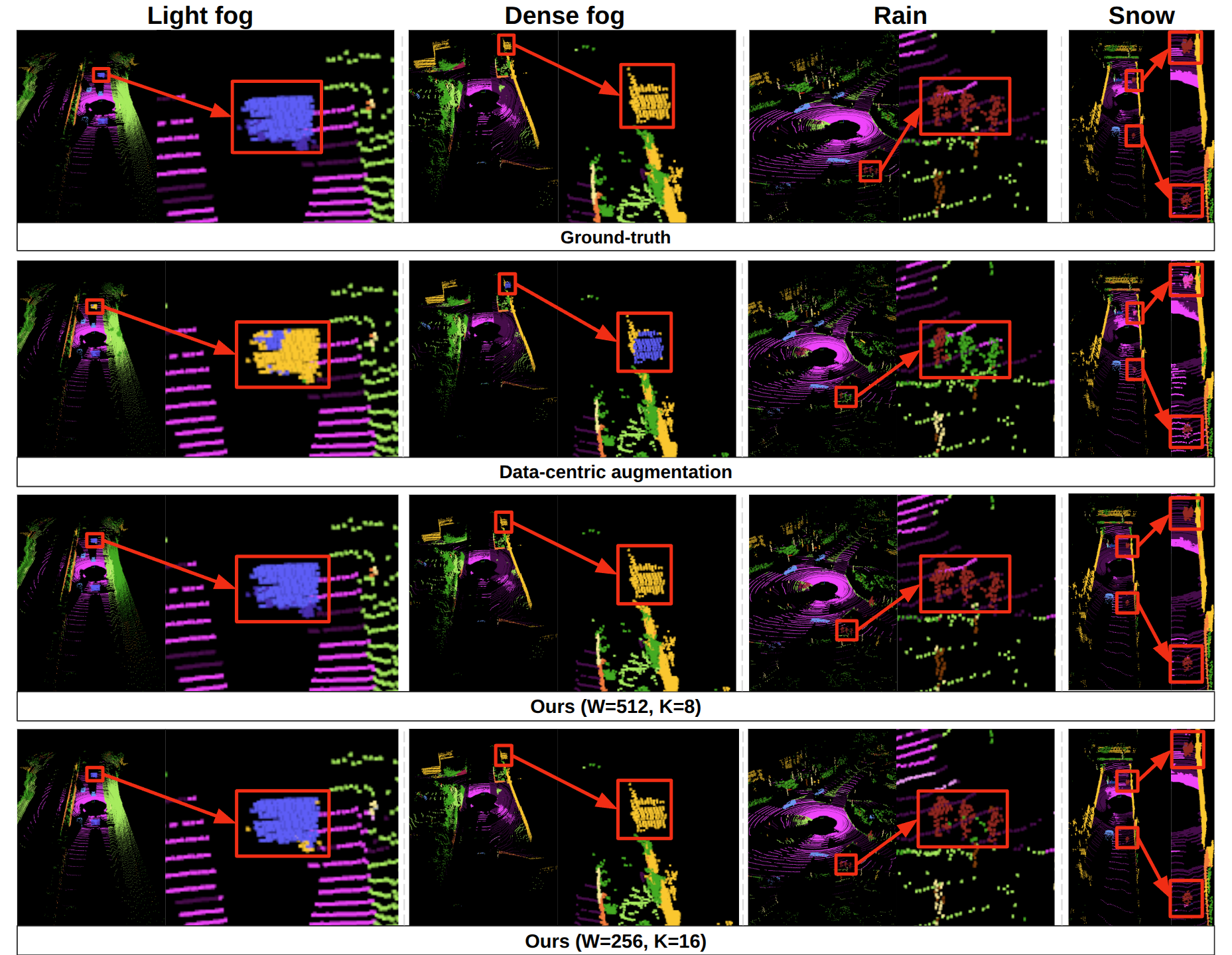}\\[4pt]
    \caption{Qualitative results on SemanticSTF~\cite{xiao2023pointdr} in \textbf{light fog}, \textbf{dense fog}, \textbf{rain}, and \textbf{snow}. Top: baseline~\cite{park2024lidarweather}. Middle/Bottom: ours \((W=512,K=8)\), \((W=256,K=16)\). Our adapter preserves continuity and reduces boundary confusion.}
  \label{fig:stf_weather}
\end{figure*}

\textbf{Base Objective:}
We compute the standard class-weighted cross-entropy loss:
\begin{equation}\label{eq:loss-ce}
\mathcal{L}_{\mathrm{CE}}(\hat Y, Y)
= -\frac{1}{N}\sum_{i=1}^{N}\sum_{c=1}^{C} w_c\, y_{i}^{(c)} \log \hat y_{i}^{(c)}.
\end{equation}
The reference branch loss is designated as $\mathcal{L}_{\mathrm{before}}=\mathcal{L}_{\mathrm{CE}}(\hat Y_{\mathrm{sj}},Y)$.

\textbf{Geometry-Aware Reweighting After Drop:}
To penalize structural degradation, the post-drop loss dynamically reweights the contribution of each point by $(1+\kappa s_i)$:
\begin{equation}\label{eq:loss-after}
\mathcal{L}_{\mathrm{after}}
= \frac{1}{N}\sum_{i=1}^{N} (1+\kappa s_i)
   \left[-\sum_{c=1}^{C} w_c\, y_{i}^{(c)} \log \hat y_{i,\mathrm{drop}}^{(c)}\right].
\end{equation}

\textbf{Entropy Regularization:}
To mitigate network overconfidence and enforce consistency between the dual branches, we incorporate an entropy term. Defining $H(\mathbf p)=-\sum_{c=1}^{C} p^{(c)}\log p^{(c)}$, the regularization loss is formulated as:
\begin{equation}\label{eq:loss-entropy}
\mathcal{L}_{\mathrm{ent}}
= \frac{1}{2N}\sum_{i=1}^{N}\Big(
H(\hat{\mathbf y}_{i,\mathrm{sj}})+H(\hat{\mathbf y}_{i,\mathrm{drop}})
\Big).
\end{equation}

\textbf{Total Optimization Objective:}
The final loss optimized during training is:
\begin{equation}\label{eq:loss-total}
\mathcal{L}_{\mathrm{seg}}
= \mathcal{L}_{\mathrm{before}} + \alpha\,\mathcal{L}_{\mathrm{after}} + \eta\,\mathcal{L}_{\mathrm{ent}}.
\end{equation}

\begin{table}[t]
\centering
\scriptsize
\setlength{\tabcolsep}{4pt}
\caption{Hyperparameter ablation of the Light Geometry-Aware Adapter (window size $W$, neighbors $K$) on SemanticKITTI$\rightarrow$SemanticSTF.}
\label{tab:ablation_window_k}
\begin{tabular}{l|cccccc|c}
\toprule
Method & car & bi.cle & mt.cle & truck & bus & pers & mIoU \\
\midrule
LiDARWeather~\cite{park2024lidarweather} & 83.1 & 1.2 & 17.2 & 30.5 & 18.4 & 47.5 & 35.8 \\
No Thing, Nothing~\cite{park2025ntn}     & \underline{83.3} & 3.7 & \textbf{31.3} & \underline{36.2} & 18.2 & \textbf{53.3} & 38.9 \\
\midrule
Ours (W=256, K=8)   & 85.6 & 7.1 & 28.0 & 31.5 & 26.3 & \textbf{51.1} & 37.8 \\
Ours (W=256, K=16)  & 85.0 & \textbf{9.9} & 23.9 & 38.7 & 23.6 & 46.3 & \textbf{39.2} \\
Ours (W=512, K=8)   & 84.5 & 8.4 & 20.5 & \textbf{41.0} & 27.3 & 48.1 & 38.0 \\
Ours (W=512, K=16)  & \textbf{86.0} & 6.8 & 19.1 & 37.7 & \textbf{30.6} & 49.5 & 37.1 \\
\bottomrule
\end{tabular}
\end{table}

\begin{table}[t]
\centering
\scriptsize
\setlength{\tabcolsep}{4pt}
\caption{Component ablation analysis on SemanticKITTI$\rightarrow$SemanticSTF. 'C' denotes Circular Padding, 'L' denotes Local-window KNN.}
\label{tab:ablation_2}
\begin{tabular}{cc|c|cccccc|c}
\toprule
\multicolumn{2}{c|}{Adapter} & \multirow{2}{*}{Drop} &
\multirow{2}{*}{car} & \multirow{2}{*}{bi.cle} & \multirow{2}{*}{mt.cle} & \multirow{2}{*}{truck} & \multirow{2}{*}{bus} & \multirow{2}{*}{pers} &
\multirow{2}{*}{mIoU} \\
C & L &  &  &  &  &  &  &  & \\
\midrule
$\checkmark$ & $\times$ & $\times$ & 85.3 & 7.8 & 17.4 & 36.4 & 22.1 & 51.3 & 38.0 \\
$\times$ & $\checkmark$ & $\times$ & 85.6 & 6.5 & 25.2 & 37.5 & 23.7 & 51.6 & 37.9 \\
$\checkmark$ & $\checkmark$ & $\times$ & \textbf{85.8} & 8.1 & \textbf{27.5} & 37.2 & 22.2 & 49.0 & 38.3 \\
\midrule
$\checkmark$ & $\times$ & $\checkmark$ & 85.5 & 9.0 & 19.7 & 34.3 & \textbf{27.5} & 48.2 & 38.4 \\
$\times$ & $\checkmark$ & $\checkmark$ & 85.4 & 10.7 & 20.6 & 38.5 & 23.7 & 51.6 & 38.0 \\
$\times$ & $\times$ & $\checkmark$ & 85.2 & \textbf{11.8} & 18.6 & 27.4 & 26.4 & \textbf{56.0} & 38.0 \\
$\checkmark$ & $\checkmark$ & $\checkmark$ & 85.0 & 9.9 & 23.9 & \textbf{38.7} & 23.6 & 46.3 & \textbf{39.2} \\
\bottomrule
\end{tabular}
\end{table}

\begin{table*}[t]
\centering
\scriptsize
\setlength{\tabcolsep}{2.5pt}
\caption{Comparison of methods on the SynLiDAR$\rightarrow$SemanticSTF benchmark.}
\label{tab:synlidar_to_stf_final}
\begin{tabular}{l|ccccccccccccccccccc|c}
\toprule
\textbf{Method} & car & bi.cle & mt.cle & truck & oth-v. & pers. & bi.clst & mt.clst & road & parki. & sidew. & othe.g. & build. & fence & veget. & trunk & terra. & pole & traf. & \textbf{mIoU} \\
\midrule
Oracle & 89.4 & 42.1 & 0.0 & 59.9 & 61.2 & 69.6 & 39.0 & 0.0 & 82.2 & 21.5 & 58.2 & 45.6 & 86.1 & 63.6 & 80.2 & 52.0 & 77.6 & 50.1 & 61.7 & 54.7 \\
Source-only & 27.1 & \underline{3.0} & 0.6 & 15.8 & 0.1 & 25.2 & 1.8 & 5.6 & 23.9 & 0.3 & 14.6 & 0.6 & 36.3 & 19.9 & 37.9 & 17.9 & 41.8 & 9.5 & 2.3 & 15.0 \\
Dropout~\cite{srivastava2014dropout} & 28.0 & \underline{3.0} & 1.4 & 9.6 & 0.0 & 17.1 & 0.8 & 0.7 & 34.2 & 6.8 & \textbf{30.5} & 1.1 & 35.5 & 19.1 & 42.3 & 17.6 & 36.0 & 14.0 & 2.8 & 15.2 \\
Perturbation & 27.1 & 2.3 & 2.3 & 16.0 & 0.1 & 23.7 & 1.2 & 4.0 & 27.0 & 3.6 & 16.2 & 0.8 & 29.2 & 16.7 & 35.3 & 18.3 & 17.9 & 5.1 & 2.4 & 15.2 \\
PolarMix~\cite{xiao2022polarmix} & 39.2 & 1.1 & 2.2 & 8.3 & 1.5 & 17.8 & 0.8 & 0.7 & 23.3 & 1.3 & 17.5 & 0.4 & 45.2 & 24.8 & 46.2 & 20.1 & 38.7 & 10.9 & 0.6 & 15.7 \\
MMD~\cite{li2018dgafl} & 25.5 & 2.3 & 2.1 & 13.2 & 0.7 & 22.1 & 1.4 & 7.5 & 30.8 & 0.4 & 17.6 & 0.4 & 30.9 & 19.7 & 37.6 & 19.3 & \underline{43.5} & 9.9 & 2.6 & 15.1 \\
PCL~\cite{yao2022pcl} & 30.9 & 0.8 & 1.4 & 10.0 & 0.4 & 23.3 & \textbf{4.0} & 7.9 & 28.5 & 1.3 & 17.7 & \underline{1.2} & 39.4 & 18.5 & 40.0 & 18.0 & 38.6 & 12.1 & 2.3 & 15.5 \\
PointDR~\cite{xiao2023pointdr} & 37.8 & 2.5 & 2.4 & \underline{23.6} & 0.1 & 26.3 & 2.2 & 7.7 & 27.9 & 7.7 & 17.5 & 0.5 & 47.6 & 25.3 & 45.7 & 21.0 & 37.5 & 17.9 & 5.5 & 18.5 \\
DGLSS~\cite{kim2023dglss} & 47.9 & 2.9 & \underline{3.4} & 17.4 & 1.1 & 28.0 & 2.4 & 7.3 & 28.8 & \underline{10.2} & 18.1 & 0.2 & 48.9 & 25.3 & 46.5 & 21.4 & \textbf{45.2} & 17.9 & 4.9 & 19.8 \\
UniMix~\cite{zhao2024unimix} & \textbf{65.4} & 0.1 & \textbf{3.9} & 16.9 & \textbf{5.3} & \textbf{32.3} & 2.0 & \textbf{19.3} & \textbf{52.1} & 5.0 & \underline{27.3} & \textbf{3.0} & 49.4 & 20.3 & \textbf{58.5} & 22.7 & 23.2 & \textbf{26.1} & \textbf{20.9} & \textbf{23.4} \\
DGUIL~\cite{he2024dguil} & 43.3 & 2.8 & 2.6 & 23.2 & \underline{3.2} & \underline{31.3} & 2.5 & 4.4 & 34.3 & 9.2 & 17.9 & 0.3 & \textbf{57.1} & \underline{27.6} & \underline{50.0} & \underline{24.2} & 41.5 & 19.0 & 6.1 & 21.1 \\
\midrule
LiDARWeather~\cite{park2024lidarweather} & 39.0 & 2.5 & 2.5 & 22.3 & 0.3 & 27.0 & 1.8 & 4.0 & 36.1 & \textbf{10.3} & 19.0 & 1.0 & 50.6 & 24.5 & 45.1 & 23.2 & 34.1 & 21.9 & 7.2 & 19.6 \\
No Thing, Nothing~\cite{park2025ntn} & 48.4 & 1.5 & 2.4 & 19.4 & 0.2 & 29.1 & \underline{3.2} & \underline{8.9} & 43.5 & 6.7 & 20.5 & 0.0 & 52.2 & \textbf{30.1} & 49.8 & 20.0 & 32.9 & \underline{24.7} & \underline{7.5} & 21.1 \\
Ours & \underline{49.1} & \textbf{5.0} & 2.3 & \textbf{26.2} & 1.8 & 29.8 & 1.9 & 6.4 & \underline{48.8} & 8.5 & 21.6 & 0.0 & \underline{55.2} & 25.5 & 48.6 & \textbf{24.3} & 40.1 & 20.0 & 6.2 & \underline{22.1} \\
\bottomrule
\end{tabular}
\end{table*}

\section{Experiments}

\subsection{Setup}
All models are evaluated under the rigorous \textbf{SemanticKITTI$\rightarrow$SemanticSTF} \emph{source-only} cross-weather paradigm. Models are trained entirely on SemanticKITTI~\cite{behley2019semantickitti} and evaluated directly on SemanticSTF~\cite{xiao2023pointdr}, strictly forbidding target-domain labels or fine-tuning. We report the mean Intersection over Union (mIoU) and per-class IoU across 19 semantic categories. To ensure statistical reliability given the competitive margins against advanced baselines, we execute our method and the class-centric baseline across three random seeds, observing highly consistent improvements in mIoU and specifically within safety-critical classes.

\subsection{Main Results and Comparative Analysis}
Table~\ref{tab:semantic_kitti_to_stf_final} summarizes the quantitative results. Our full method achieves \textbf{39.2} mIoU, outperforming strong data-centric baselines and reaching performance comparable to the advanced class-centric regularization framework~\cite{park2025ntn}. Fig.~\ref{fig:stf_weather} further shows qualitative results across light fog, dense fog, rain, and snow, where our approach produces sharper boundaries and fewer label inversions in structurally sparse regions.

\textbf{Mechanisms of the Adapter and Region-Level Drop:}
The Light Geometry-Aware Adapter stabilizes learning under weather-induced sparsity by preserving \emph{neighbor continuity} and exposing \emph{structural risks} at boundaries and corners, thereby suppressing boundary fragmentation and local class inversions. Region-level Drop complements this by perturbing \emph{structurally coherent regions} instead of isolated points, reducing supervision damage near object perimeters. Together, geometry-conditioned dropping yields improved robustness, particularly for large rigid and safety-critical categories.

\textbf{Comparison to Class-Centric Regularization:}
Class-centric regularization directly enforces class-wise priors, providing notable advantages for rare or structurally thin categories that suffer extreme degradation in adverse weather. Our methodology, by contrast, is intrinsically neighborhood-driven. Thus, its efficacy is maximized when localized structural support remains partially intact. When point clouds become severely decimated, class-centric priors may retain an edge; however, our geometry-driven approach provides a fundamentally orthogonal and highly competitive signal, as evidenced by our superior performance across dominant structural categories.

\textbf{Limitations and Failure Cases:}
Failures persist when local evidence is insufficient for stable region formation and cue aggregation.
(i) \emph{Extreme long-range sparsity:} Dense fog or snow yields few returns on thin structures, making $F_k$ noisy and degrading geometrically thin classes (e.g., poles, traffic signs).
(ii) \emph{Severe occlusion/truncation:} Fragmented visibility near depth discontinuities can merge distinct surfaces, inducing region leakage and boundary label bleeding.
(iii) \emph{Missing points:} Large spatial voids bias pooling toward dominant background context, and the RL drop may over-perturb fragile regions.
Finally, $(W,K)$ trades sparsity tolerance against boundary sharpness; the full adapter is generally the most reliable.
Future work will explore uncertainty-adaptive neighborhoods and occlusion-aware region partitioning to mitigate these cases.

\subsection{Ablation and Generalization}
\textbf{Ablation.}
Table~\ref{tab:ablation_window_k} studies the hyperparameters ($W$, $K$).
The adapter consistently improves over the data-centric baseline, with the best setting reaching \textbf{39.2} mIoU.
Larger $K$ helps aggregate large rigid objects but can smooth fine boundaries, while larger $W$ increases candidates under sparsity yet may introduce irrelevant neighbors when structure is fragmented. Table~\ref{tab:ablation_2} isolates components and their interaction with Region-level Drop.
Circular padding (C) stabilizes continuity at the azimuth seam, and local-window KNN (L) suppresses spurious long-range matches.
Their combination yields the most reliable cues, and Region-level Drop performs best with the full adapter by forming coherent regions without fragmenting boundaries.

\textbf{Generalization.}
To test robustness beyond a single source dataset, we additionally evaluate SynLiDAR$\rightarrow$SemanticSTF.
As shown in Table~\ref{tab:synlidar_to_stf_final}, our method remains competitive under the synthetic-to-real shift, achieving \textbf{22.1} mIoU.
Importantly, the qualitative trend is consistent with the main transfer: geometry-aware cues primarily help classes where local continuity and region consistency are informative (e.g., large rigid and safety-critical categories), while extremely sparse or occluded structures remain challenging.
This cross-source result supports that the proposed geometry-aware cues and region-level dropping are not tied to a specific source domain and can preserve robustness under substantial distribution changes.

\section{CONCLUSIONS}
In this paper, we introduced the \textbf{Light Geometry-Aware Adapter}, a novel mechanism that drastically enhances region-level point dropping for LiDAR semantic segmentation in adverse weather. Evaluated on the rigorous SemanticKITTI$\rightarrow$SemanticSTF benchmark, our approach boosts the mIoU by \textbf{+3.4} over robust data-centric baselines, achieving performance on par with advanced class-centric regularization models. The adapter is fundamentally lightweight, highly computationally efficient, and strictly plug-and-play during training. Future research will focus on integrating these geometric abstractions with dedicated, physics-based adverse-weather simulators to further bridge the sim-to-real domain gap.

\addtolength{\textheight}{-12cm}   


\IEEEtriggeratref{17}                 
\IEEEtriggercmd{\enlargethispage{\baselineskip}}

\end{document}